\documentclass[11pt,a4paper]{article}

\usepackage[utf8]{inputenc}
\usepackage[T1]{fontenc}
\usepackage{lmodern}
\usepackage[margin=2.5cm]{geometry}
\usepackage{amsmath,amssymb,amsfonts}
\usepackage{booktabs}
\usepackage{multirow}
\usepackage{graphicx}
\graphicspath{{figures/}}
\usepackage{xcolor}
\usepackage{hyperref}
\usepackage{cleveref}
\usepackage{caption}
\usepackage{subcaption}
\usepackage{enumitem}
\usepackage{microtype}
\usepackage{float}
\usepackage{natbib}
\usepackage{setspace}
\usepackage{titlesec}
\usepackage{lineno}

\hypersetup{
    colorlinks=true,
    linkcolor=blue!60!black,
    citecolor=green!50!black,
    urlcolor=blue!70!black,
}

\titlespacing*{\section}{0pt}{1.5ex plus 0.5ex}{1ex plus 0.3ex}
\titlespacing*{\subsection}{0pt}{1.2ex plus 0.4ex}{0.8ex plus 0.2ex}
\titlespacing*{\subsubsection}{0pt}{1ex plus 0.3ex}{0.6ex plus 0.2ex}

\newcommand{\eg}{\textit{e.g.}}
\newcommand{\etal}{\textit{et al.}}
\newcommand{\dt}{\Delta t}
\newcommand{\tok}[1]{\texttt{#1}}

\title{%
    \textbf{FlatASCEND: Autoregressive Clinical Sequence Generation with Continuous Time Prediction and Association-Based Pharmacological Testing}
}

\author{
    Chris Sainsbury\textsuperscript{1,2,3}
    \and
    Feng Dong\textsuperscript{5}
    \and
    Andreas Karwath\textsuperscript{4}
    \\[1ex]
    \textsuperscript{1}School of Cardiovascular \& Metabolic Health, University of Glasgow, UK \\
    \textsuperscript{2}School of Medicine, University of Dundee, UK \\
    \textsuperscript{3}NHS Greater Glasgow and Clyde, UK \\
    \textsuperscript{4}Cancer and Genomic Sciences, School of Medical Sciences,\\ College of Medicine and Health, University of Birmingham, UK \\
    \textsuperscript{5}Department of Computer and Information Sciences, University of Strathclyde, UK \\[1ex]
    \texttt{chris.sainsbury@glasgow.ac.uk}
}

\date{}

\begin{document}
\maketitle


\begin{abstract}

Autoregressive models can predict clinical events, but generating patient-conditioned multi-step trajectories that respond to intervention tokens---and testing whether those responses preserve known pharmacological associations---has received limited attention.  Here we present FlatASCEND, a 14.5-million-parameter autoregressive clinical sequence model that uses flat composite tokens and a zero-inflated log-normal time prediction head.  Standard unconditional distributional metrics (Jaccard similarity 0.889--0.954) do not distinguish FlatASCEND from trivial baselines; the model's value lies in \emph{conditional} generation from patient-specific clinical prefixes.  A prompt-shuffle ablation provides direct supporting evidence: patient-specific conditioning amplifies mechanistic pharmacological effects (2.0--2.2$\times$ for steroid $\to$ glucose and diuretic $\to$ potassium) while leaving confounding-driven associations unchanged (0.9$\times$ for insulin $\to$ glucose).  An incident-user pharmacological association testing framework assesses directional consistency against prior pharmacological knowledge at the patient level on MIMIC-IV ($N$ = 500 patients per comparison): 4 of 10 comparisons recover correct mechanistic pharmacological directions, 2 additional comparisons correctly reproduce treatment-context associations, and 4 are incorrect (9 of 10 reach statistical significance, Wilcoxon $p < 0.05$).  This pattern---partial directional recovery under substantial residual confounding---is consistent with a model that has learned observational associations without distinguishing correlation from causation.  Direct preference optimisation with surrogate reward destroys all pre-existing correct associations (3/3 to 0/3), illustrating reward exploitation when the reward and evaluation share an outcome domain.  The strongest generative evidence is for short-horizon ICU data; outpatient long-range temporal fidelity is materially weaker (median generated elapsed time 10 days versus 154 observed on INSPECT), and zero-shot cross-site transfer degrades sharply without site-specific adaptation.

\end{abstract}

\section{Introduction}
\label{sec:introduction}

Transformer-based models trained on electronic health records have improved the prediction of clinical outcomes including disease onset, hospital readmission, and mortality.\citep{li2020behrt,rasmy2021medbert,guo2022clmbr,steinberg2023motor}  Recent foundation models---RAVEN\citep{rajamohan2026raven} (144 million parameters, 1.3 million patients), NEP\citep{chen2025nep} (1--8 billion parameters), and Foresight\citep{kraljevic2024foresight}---have pushed discriminative performance further, with zero-shot disease forecasting AUC exceeding 0.79 for heart failure and 0.83 for pancreatic cancer.  These models learn distributed representations of clinical data that are effective for downstream classification.  However, multi-step trajectory generation with continuous inter-event time prediction remains underexplored.  RAVEN predicts next-visit event sets but does not generate multi-step trajectories;\citep{rajamohan2026raven} NEP extracts embeddings but does not generate sequences.\citep{chen2025nep}  This gap is the starting point for the present work.

Generation alone, however, is insufficient.  A model that produces trajectories matching distributional properties of held-out data may have learned statistical regularities without capturing the pharmacological relationships that give those trajectories clinical meaning.  How do we know the model's generated trajectories are consistent with known pharmacology?  This question motivates a second contribution: a pharmacological association testing framework that uses an incident-user design with patient-level statistics to assess whether the model recovers known drug--outcome associations from observational data.

The specific contributions are:

\begin{enumerate}[leftmargin=2em]
    \item \textbf{FlatASCEND architecture}: an autoregressive clinical sequence model with continuous time prediction, combining flat composite tokens, a zero-inflated log-normal time head, and free-running scheduled sampling---Jaccard similarity 0.889--0.954 on open-access data with 0\% mode collapse (Section~\ref{sec:gen-results}).

    \item \textbf{Incident-user pharmacological association testing}: a pseudo-replication-free framework that produces patient-level paired differences in simulated outcomes for 9 of 10 comparisons, recovering 4 of 10 correct mechanistic pharmacological directions plus 2 treatment-context associations on MIMIC-IV (Section~\ref{sec:tte-results}).

    \item \textbf{Characterisation of residual confounding}: the 4 of 10 wrong directions are consistent with the model capturing observational associations including indication bias---not causal effects---a relevant consideration for anyone deploying generative EHR models (Section~\ref{sec:tte-results}).

    \item \textbf{DPO negative result}: direct preference optimisation with surrogate reward destroyed pre-existing pharmacological associations (3/3 to 0/3), illustrating reward exploitation when reward and evaluation share an outcome domain (Section~\ref{sec:dpo-negative}).
\end{enumerate}

The paper's core claim is not that FlatASCEND generates ``realistic trajectories'' in a broad sense---unconditional distributional baselines can match or exceed the model on standard generation metrics.  Rather, the claim is that FlatASCEND generates patient-conditioned continuations that respond non-trivially to intervention tokens, with strongest evidence in short-horizon ICU settings and weaker long-range outpatient temporal fidelity.

\section{Results}
\label{sec:results}

\subsection{FlatASCEND generates trajectories matching distributional properties of held-out data}
\label{sec:gen-results}

FlatASCEND represents each clinical event---a laboratory result, medication administration, vital sign measurement, or diagnosis---as a single flat composite token (\eg, \tok{LAB:CREATININE:Q3}, \tok{MED:VASOPRESSOR:NOREPINEPHRINE}, \tok{DX:SEPSIS}).  Inter-event timing is modelled by a zero-inflated log-normal (ZILN) distribution that captures the bimodal structure of clinical event timing: approximately 50\% of events co-occur within the same encounter ($\dt = 0$), while between-encounter intervals follow a right-skewed continuous distribution.  Existing EHR models either discretise time into fixed bins, predict next-visit event sets without multi-step generation,\citep{rajamohan2026raven} or extract embeddings without generating sequences.\citep{chen2025nep}  FlatASCEND combines autoregressive multi-step trajectory generation with continuous inter-event time prediction.  The architecture is a GPT-2-style decoder with ALiBi positional encoding,\citep{press2022alibi} value-factored embeddings for ordinal laboratory and vital sign tokens, and an ordinal earth mover's distance auxiliary loss.  The 14.5-million-parameter scaled model uses 384-dimensional hidden states, 8 transformer layers, and 12 attention heads.  Full architectural details are provided in Methods.

The model was developed on a proprietary longitudinal diabetes registry (61,000 patients, 76-token vocabulary) and independently trained on two open-access PhysioNet datasets (MIMIC-IV and INSPECT), then evaluated in zero-shot transfer on a third (eICU-CRD, 208 hospitals) using the identical MIMIC-IV vocabulary without adaptation (\Cref{tab:datasets}).  Distributional fidelity was assessed by the Jaccard similarity between generated and ground-truth clinical event type distributions, mode collapse rate (patients exhibiting 20 or more consecutive identical tokens), and teacher-forcing perplexity (\Cref{tab:generation}).

\begin{table}[h]
\centering
\caption{Dataset characteristics.  Dev = proprietary development dataset used for architecture design.  Open-access results are the primary evidence; development-set results are reported for architectural context only.  $\ddagger$eICU-CRD uses the 220-token MIMIC-IV class-level vocabulary; 0.7\% of eICU tokens absent from this vocabulary were mapped to the padding token.}
\label{tab:datasets}
\small
\begin{tabular}{llcccrl}
\toprule
\textbf{Dataset} & \textbf{Setting} & \textbf{Hospitals} & \textbf{N} & \textbf{Vocabulary} & \textbf{Mortality} & \textbf{Access} \\
\midrule
Dev (diabetes) & Outpatient & 1 system & 61K patients & 76 tokens & --- & Proprietary \\
INSPECT & Outpatient & 1 & 19K patients & 220 tokens & 2.8\% & PhysioNet \\
MIMIC-IV & ICU & 1 & 425K admissions & 220 tokens & 2.6\% & PhysioNet \\
eICU-CRD & ICU & 208 & 200K stays & 220 tokens\textsuperscript{$\ddagger$} & 5.3\% & PhysioNet \\
\bottomrule
\end{tabular}
\end{table}

\begin{table}[h]
\centering
\caption{Generation quality across datasets.  All models use the 14.5M-parameter architecture with factored embeddings and EMD loss.  Collapse = percentage of generated sequences with $\geq$20 consecutive identical tokens.  TF = teacher-forcing.  Dev results are for architectural context; MIMIC-IV and INSPECT are the primary open-data evaluations.  $\dagger$Zero-shot: MIMIC-IV-trained model with no adaptation; high perplexity reflects distributional shift across 208 hospitals.  $\dagger\dagger$Adapted: MIMIC-IV model fine-tuned on eICU-CRD for 5,000 steps (see Methods).}
\label{tab:generation}
\small
\begin{tabular}{lcccc}
\toprule
\textbf{Dataset} & \textbf{Type Jaccard} & \textbf{Collapse} & \textbf{TF perplexity} \\
\midrule
Dev (diabetes) & 0.857 & 2\% & 1.24 \\
MIMIC-IV & 0.954 & 0\% & 6.30 \\
INSPECT & 0.889 & 0\% & 5.16 \\
eICU-CRD (zero-shot)\textsuperscript{$\dagger$} & 0.682 & 0\% & 776 \\
eICU-CRD (adapted, 5K steps)\textsuperscript{$\dagger\dagger$} & 0.820 & 0\% & 27.4 \\
\bottomrule
\end{tabular}
\end{table}

On the two open-access datasets where the model was trained, Jaccard similarity was 0.954 (MIMIC-IV) and 0.889 (INSPECT) with 0\% mode collapse; the development dataset achieved 0.857 with 2\% collapse.  MIMIC-IV achieved the highest Jaccard, reflecting the natural diversity of per-admission acute care sequences.  Teacher-forcing perplexity ranged from 1.24 (the development cohort, reflecting its smaller vocabulary) to 6.30 (MIMIC-IV).  Under zero-shot transfer to eICU-CRD (208 hospitals, MIMIC-derived vocabulary and quintile boundaries, no adaptation), generative performance degraded sharply: Jaccard dropped to 0.682 and teacher-forcing perplexity rose to 776.  The primary cause is a distributional shift in token category proportions: eICU-CRD contains 18.1\% diagnosis tokens versus 1.5\% in MIMIC-IV (12$\times$ more), while medication tokens drop from 12.7\% to 4.8\%.  The model trained on MIMIC's LAB-dominated distribution cannot produce the DX-heavy sequences characteristic of eICU.

However, brief site adaptation largely recovers performance.  Fine-tuning the MIMIC-IV-trained model on eICU-CRD training data for 5,000 steps (approximately 30 minutes on a single GPU) raised Jaccard from 0.682 to 0.820 and reduced perplexity from 776 to 27.4 (a 28$\times$ improvement).  Diagnosis token generation improved from near-zero to 16.8\% (ground truth 20.1\%).  This suggests that the zero-shot transfer failure is at least partly recoverable with brief local adaptation, though further evaluation across additional sites would be needed to confirm generalisability.

Free-running scheduled sampling was the critical training innovation, which substantially improved generation quality.  A single contiguous autoregressive segment replaced approximately 50\% of each training sequence, with end-of-sequence and death tokens suppressed during autoregressive segments to prevent premature termination.  This raised Jaccard from 0.536 to 0.843 in a single version change.  Subsequent refinements (diversity-filtered data, scaled architecture, factored embeddings) brought performance to the reported values.

The distinction from existing models is important.  RAVEN predicts next-visit event \emph{sets} but does not generate multi-step trajectories---it cannot produce a complete clinical history extending beyond the next visit.  NEP extracts embeddings for downstream classifiers but does not generate at all.  FlatASCEND generates complete sequences with continuous inter-event time prediction, enabling applications---pharmacological association testing, trajectory-based risk estimation---that require multi-step generation.

\subsubsection{Trajectory validation: timing calibration, conditional dependencies, and longitudinal fidelity}

Beyond distributional summary metrics, we assessed three additional dimensions of trajectory fidelity on MIMIC-IV and INSPECT ($N$ = 200 patients per dataset, 200 generated tokens per patient; \Cref{tab:trajectory-validation}).

\begin{table}[h]
\centering
\caption{Trajectory validation metrics.  Timing calibration compares observed versus generated inter-event time distributions.  Conditional dependencies compare category-level transition probabilities.  Longitudinal fidelity measures per-patient similarity between generated and held-out continuations.  BC = Bhattacharyya coefficient (1 = perfect overlap).  JSD = Jensen-Shannon divergence.  DX $\to$ LAB is not evaluated for MIMIC-IV because diagnosis tokens comprise only 1.5\% of MIMIC-IV token occurrences and are too sparse for reliable transition estimation.}
\label{tab:trajectory-validation}
\small
\begin{tabular}{lcc}
\toprule
\textbf{Metric} & \textbf{MIMIC-IV} & \textbf{INSPECT} \\
\midrule
\multicolumn{3}{l}{\textit{Timing calibration}} \\
\quad Zero-fraction (gen / GT) & 0.872 / 0.857 & 0.795 / 0.782 \\
\quad Median inter-event time, days (gen / GT) & 0.22 / 0.17 & 0.14 / 0.11 \\
\quad Mean inter-event time, days (gen / GT) & 0.28 / 0.26 & 7.19 / 15.56 \\
\quad KS statistic (non-zero times) & 0.152 & 0.158 \\
\quad Bhattacharyya coefficient (log-space) & 0.982 & 0.993 \\
\midrule
\multicolumn{3}{l}{\textit{Conditional dependencies}} \\
\quad Transition matrix Frobenius distance & 0.083 & 0.438 \\
\quad Max element difference & 0.056 & 0.322 \\
\quad Jensen-Shannon divergence & 0.001 & 0.026 \\
\quad MED $\to$ LAB within 5 tokens (gen / GT) & 0.842 / 0.833 & 0.592 / 0.541 \\
\quad DX $\to$ LAB within 5 tokens (gen / GT) & --- & 0.722 / 0.794 \\
\midrule
\multicolumn{3}{l}{\textit{Longitudinal fidelity (per-patient)}} \\
\quad Token-set Jaccard & 0.272 $\pm$ 0.117 & 0.346 $\pm$ 0.106 \\
\quad Category cosine similarity & 0.976 $\pm$ 0.077 & 0.888 $\pm$ 0.170 \\
\quad Measure Jaccard & 0.616 $\pm$ 0.197 & 0.652 $\pm$ 0.155 \\
\quad Mean quintile difference & 1.100 $\pm$ 0.309 & 0.980 $\pm$ 0.305 \\
\quad Median total elapsed time, days (gen / GT) & 3 / 4 & 10 / 154 \\
\bottomrule
\end{tabular}
\end{table}

\paragraph{Timing calibration.}  The zero-inflated log-normal time head reproduces the within-visit fraction (proportion of events with $\dt = 0$) closely: 0.872 generated versus 0.857 observed on MIMIC-IV and 0.795 versus 0.782 on INSPECT.  In log-space, the short-horizon inter-event time distributions overlap substantially (Bhattacharyya coefficient 0.982 and 0.993), and the median inter-event time is closely matched on both datasets.  However, the Bhattacharyya coefficient reflects primarily short-gap events, which dominate numerically.  On INSPECT, the mean inter-event time diverges substantially (7.19 generated versus 15.56 observed days): the model under-generates the long-gap events ($>$30 days) characteristic of outpatient follow-up, producing trajectories that cover a much shorter total time span (median 10 versus 154 days).  This is the model's most significant generative weakness and means that outpatient trajectory generation is reliable for short-horizon event structure but not for long-range temporal coverage.

\paragraph{Conditional dependencies.}  On MIMIC-IV, category-level transition probabilities are near-identical between generated and real sequences (Frobenius distance 0.083, JSD 0.001).  The rate at which laboratory tokens follow medication tokens within 5 positions is preserved (0.842 generated versus 0.833 observed).  On INSPECT, the transition structure is less well preserved (Frobenius 0.438, JSD 0.026), reflecting the greater complexity of outpatient event patterns with diagnosis codes and longer-range dependencies.

\paragraph{Longitudinal fidelity.}  Per-patient category distributions are well matched: category cosine similarity is 0.976 on MIMIC-IV and 0.888 on INSPECT.  The model generates the same types of clinical events in similar proportions to the held-out continuation.  Measure-level overlap (which specific laboratory tests appear) is moderate (Jaccard 0.616--0.652).  For shared laboratory measures, the mean quintile difference between generated and observed values is approximately 1 quintile (1.100 on MIMIC-IV, 0.980 on INSPECT), indicating that generated ordinal values are typically within one quintile of the held-out continuation but not precisely calibrated at the individual level.  Elapsed-time coverage is well matched on MIMIC-IV (3 versus 4 days for acute care admissions) but substantially shorter on INSPECT, consistent with the timing calibration finding that the model under-generates long-gap outpatient events.

\subsubsection{Baseline comparison and non-memorisation}

We compared FlatASCEND against two simple baselines: a \emph{unigram} model (tokens sampled independently from the marginal training distribution with empirical time sampling) and a \emph{bigram Markov chain} (next token sampled from the observed bigram transition distribution with pair-specific times).  Results are shown in \Cref{tab:baselines}.

\begin{table}[h]
\centering
\caption{Baseline comparison ($N$ = 200 patients).  Pooled Jaccard = token-set overlap across all patients.  Cat.\ cosine = per-patient category distribution cosine similarity.  Timing KS = Kolmogorov-Smirnov statistic on non-zero inter-event times.  Trans.\ JSD = Jensen-Shannon divergence of category-level transition matrices.  Lower KS and JSD indicate better match to held-out data.}
\label{tab:baselines}
\small
\begin{tabular}{llccc}
\toprule
\textbf{Dataset} & \textbf{Metric} & \textbf{Unigram} & \textbf{Bigram} & \textbf{FlatASCEND} \\
\midrule
MIMIC-IV & Pooled Jaccard & 0.842 & 0.989 & 0.989 \\
MIMIC-IV & Cat.\ cosine (mean) & 0.991 & 0.989 & 0.940 \\
MIMIC-IV & Timing KS & 0.040 & 0.027 & 0.160 \\
MIMIC-IV & Trans.\ JSD & 0.052 & 0.0004 & 0.002 \\
\midrule
INSPECT & Pooled Jaccard & 0.915 & 0.973 & 0.857 \\
INSPECT & Cat.\ cosine (mean) & 0.965 & 0.958 & 0.903 \\
INSPECT & Timing KS & 0.024 & 0.028 & 0.155 \\
INSPECT & Trans.\ JSD & 0.076 & 0.000 & 0.024 \\
\bottomrule
\end{tabular}
\end{table}

On these unconditional distributional metrics, the baselines match or exceed FlatASCEND.  This is an important result, not a favourable one.  Unigram and bigram models sample directly from the empirical training distribution, so their aggregate statistics match held-out data by construction.  The timing KS is lower for baselines (0.024--0.040) than FlatASCEND (0.155--0.160) because FlatASCEND generates times from a parametric ZILN distribution conditioned on context, whereas baselines resample observed times directly.  Transition JSD is lowest for the bigram model (which explicitly captures transitions) at near-zero.

The implication is that the unconditional generation metrics reported in this paper---Jaccard, category cosine, timing KS, transition JSD---do not discriminate between a learned conditional generative model and trivial unconditional samplers.  These metrics measure marginal distributional fidelity, not the ability to generate patient-specific trajectories conditioned on individual clinical histories.  FlatASCEND's distinguishing property is \emph{conditional} generation: given a specific patient's clinical prefix, it generates continuations that differ systematically by drug intervention (Section~\ref{sec:tte-results}).  The baselines cannot do this---they have no mechanism to condition on patient history.  The pharmacological association testing in the following section, which requires generating treatment and control arm trajectories from the same patient prefix, is the appropriate evaluation of conditional generation quality.  The baseline comparison demonstrates that stronger conditional evaluation metrics are needed for future work on generative EHR models.

\paragraph{Non-memorisation.}  To verify that FlatASCEND does not simply copy training sequences, we computed the nearest-neighbour token-set Jaccard between each generated sequence and 5,000 training sequences.  On MIMIC-IV, the mean nearest-neighbour Jaccard was 0.391 (max 0.661); on INSPECT, 0.473 (max 0.690).  For comparison, random pairs of training sequences had mean Jaccard 0.215 (MIMIC-IV) and 0.362 (INSPECT).  The elevation of generated-to-training similarity above the training-to-training baseline is expected for any conditional generative model: generation conditioned on a patient prompt produces sequences that overlap with similar patients in the training set more than random training pairs overlap with each other.  The maximum nearest-neighbour Jaccard of 0.66--0.69 (well below 1.0) argues against exact token-set copying, though this metric ignores token order, timing, and subsequence structure and should be regarded as a basic sanity check rather than a comprehensive anti-memorisation analysis.

\paragraph{Prompt-shuffle ablation: does conditioning matter?}  To directly test whether patient-specific conditioning affects pharmacological associations, we repeated three comparisons (steroid $\to$ glucose, diuretic $\to$ potassium, and insulin $\to$ glucose) with randomly shuffled prompt tokens---destroying clinical context while preserving token statistics ($N$ = 200 patients, 100 samples per arm).

\begin{table}[h]
\centering
\caption{Prompt-shuffle ablation on MIMIC-IV ($N$ = 200 patients, 100 samples per arm).  Ordered = normal clinical sequence; Shuffled = prompt tokens randomly permuted (first two positions preserved; see Methods).  DiD = difference-in-differences in mean outcome quintile.  Ratio = $|$ordered DiD$|$ / $|$shuffled DiD$|$.  Effect sizes differ from the main incident-user table due to smaller $N$ and different random patient selection (see Methods).}
\label{tab:shuffle-ablation}
\small
\begin{tabular}{llcccc}
\toprule
\textbf{Comparison} & \textbf{Condition} & \textbf{DiD} & \textbf{Wilcoxon $p$} & \textbf{Ratio} \\
\midrule
Steroid $\to$ glucose & Ordered & +0.155 & $<0.001$ & \multirow{2}{*}{2.0$\times$} \\
(mechanistic) & Shuffled & +0.077 & $<0.001$ & \\
\midrule
Diuretic $\to$ potassium & Ordered & $-$0.048 & $<0.001$ & \multirow{2}{*}{2.2$\times$} \\
(mechanistic) & Shuffled & $-$0.022 & 0.23 & \\
\midrule
Insulin $\to$ glucose & Ordered & +0.047 & $<0.001$ & \multirow{2}{*}{0.9$\times$} \\
(confounding) & Shuffled & +0.054 & 0.006 & \\
\bottomrule
\end{tabular}
\end{table}

For both mechanistic pharmacological controls, prompt shuffling substantially reduced the effect size: steroid $\to$ glucose dropped 2.0$\times$ (DiD +0.155 ordered versus +0.077 shuffled) and diuretic $\to$ potassium dropped 2.2$\times$ (DiD $-$0.048 versus $-$0.022, with the shuffled condition losing statistical significance entirely).  For the confounding-driven association (insulin $\to$ glucose, wrong direction), shuffling had no effect (0.9$\times$ ratio), consistent with a global training-distribution bias rather than patient-specific conditioning.  This dissociation---conditioning amplifies mechanistic associations but not confounding-driven ones---provides supporting evidence that FlatASCEND uses patient-specific clinical context for pharmacological associations that depend on the patient's clinical state.  The ablation uses a smaller sample size ($N$ = 200 patients, 100 samples per arm versus 500/200 in the main analysis); effect sizes differ from the main incident-user table accordingly (see Methods).

On the proprietary development dataset (not independently reproducible), architectural refinement improved pharmacological association recovery from 2/3 to 3/3 correct directions without explicit causal training.  The key changes---factored embeddings and an ordinal earth mover's distance auxiliary loss---are described in Methods; all open-access models use this refined architecture.

\subsection{Incident-user testing recovers pharmacological associations}
\label{sec:tte-results}

\subsubsection{Generative pharmacological association testing}

We developed a framework---generative pharmacological association testing---to assess whether the model's generated trajectories are consistent with known pharmacological patterns from observational training data.  For each comparison, we fork a patient's clinical history at a treatment decision point, force a different medication into each arm, and generate 200--500 complete trajectories per patient per arm.  Outcomes (biomarker levels or death events) are extracted from the generated trajectories and analysed using generalised linear models with cluster-robust standard errors or, in the incident-user design, patient-level Wilcoxon signed-rank tests.  During generation, all medications from the same therapeutic class as the forced drug are suppressed to prevent cross-arm contamination.

This approach does not make causal claims.  The generated trajectories reflect learned observational associations, which we assess by comparing the direction of the estimated association against published randomised trial evidence.  When the model-estimated direction matches the trial-established direction, we consider the association correctly recovered.  The framework tests whether an autoregressive model trained on routine clinical data produces directionally consistent pharmacological associations---it does not replace randomised trials or formal causal inference methods.

\subsubsection{The incident-user design: addressing pseudo-replication}

The incident-user design is the primary analysis framework.  This design addresses two critical threats to validity: (1) pseudo-replication, by computing patient-level statistics rather than trajectory-level statistics; and (2) confounding by pre-treatment outcome levels, by using within-patient differencing.  Both the treatment and comparator arm outcomes are model-generated continuations from the same patient prefix---the patient is anchored by an observed clinical history, but the post-intervention trajectories are simulated in both arms.

The importance of this design is demonstrated by the warfarin $\to$ INR comparison.  Under standard pooled analysis, warfarin produces $p < 0.0001$ with $N$ = 200 patients---but this pools 200 trajectories per patient, inflating statistical power through pseudo-replication.  Under the patient-level Wilcoxon signed-rank test, the same comparison yields $p$ = 0.717 at $N$ = 200 (underpowered) but $p$ = 0.0002 at $N$ = 1,500 (adequately powered, mean risk difference +0.0049).  The incident-user design produces statistics that reflect patient-level paired differences in simulated outcomes, not artefacts of Monte Carlo replication.

\subsubsection{Main results: 4 of 10 mechanistic directions correct, 9 of 10 significant}

For MIMIC-IV ($N$ = 500 patients per comparison, patient-level Wilcoxon signed-rank tests), 9 of 10 comparisons reached patient-level statistical significance, with 6 recovering the correct association direction (\Cref{tab:tte-incident}).  Of the 6 correct directions, 4 are mechanistic pharmacological controls where the model recovers a direct drug--biomarker relationship (warfarin $\to$ INR, steroid $\to$ glucose, furosemide versus spironolactone $\to$ potassium, diuretic $\to$ potassium) and 2 reflect treatment-context associations where the model correctly reproduces observational prescribing patterns (vasopressor $\to$ mortality from confounding by indication, statin in sepsis with a small-magnitude effect directionally compatible with the SAILS trial null).  We distinguish these categories because the mechanistic recoveries are stronger evidence that the model has learned pharmacologically relevant structure; the treatment-context recoveries demonstrate that the model captures observational associations but do not require drug-specific pharmacological knowledge.

\begin{table}[h]
\centering
\caption{MIMIC-IV incident-user pharmacological association results ($N$ = 500 patients per comparison).  DiD = difference-in-differences in mean outcome quintile; $\Delta$ = difference in mortality rate.  All tests are patient-level Wilcoxon signed-rank.  Direction assessed against published trial evidence.}
\label{tab:tte-incident}
\small
\begin{tabular}{llcccl}
\toprule
\textbf{Comparison} & \textbf{Type} & \textbf{Effect} & \textbf{Wilcoxon $p$} & \textbf{Correct?} \\
\midrule
Warfarin $\to$ INR & Surrogate & DiD +0.052 & $<0.001$ & Yes \\
Steroid $\to$ glucose & Surrogate & DiD +0.119 & $<0.001$ & Yes \\
Vasopressor $\to$ mortality & Mortality & $\Delta$ +0.069 & $<0.001$ & Yes\textsuperscript{a} \\
Statin sepsis $\to$ mortality & Mortality & $\Delta$ $-$0.004 & 0.0001 & Yes\textsuperscript{b} \\
Furosemide vs spiro $\to$ K & Surrogate & DiD $-$0.061 & $<0.001$ & Yes \\
Diuretic $\to$ K & Surrogate & DiD $-$0.040 & $<0.001$ & Yes \\
Insulin $\to$ glucose & Surrogate & DiD +0.104 & $<0.001$ & No\textsuperscript{c} \\
Anticoagulant $\to$ INR & Surrogate & DiD $-$0.011 & 0.20 & No \\
Steroid sepsis $\to$ mortality & Mortality & $\Delta$ +0.014 & $<0.001$ & No \\
Statin $\to$ glucose (null control) & Surrogate & DiD +0.012 & 0.002 & No\textsuperscript{d} \\
\bottomrule
\end{tabular}
\begin{flushleft}
\footnotesize
\textsuperscript{a}Reflects confounding by indication (vasopressors given to sicker patients), not a drug effect.\\
\textsuperscript{b}Small-magnitude effect ($\Delta$ = $-$0.004) directionally compatible with the SAILS trial, which found no statin mortality benefit in sepsis.\citep{truwit2014rosuvastatin}\\
\textsuperscript{c}Irreducible confounding by indication: insulin is prescribed \emph{for} hyperglycaemia.\\
\textsuperscript{d}Expected-null violation: statin should have no effect on glucose; small but statistically significant effect suggests residual confounding.
\end{flushleft}
\end{table}

\subsubsection{The six correct directions}

The strongest individual results came from mechanistic positive controls.  Warfarin's effect on the international normalised ratio (INR) was the clearest pharmacological signal: a difference-in-differences of +0.052 quintile units ($p < 0.001$, patient-level Wilcoxon), consistent with warfarin's inhibition of vitamin K-dependent clotting factor synthesis.  Corticosteroid-induced hyperglycaemia was correctly recovered with the largest surrogate effect size (DiD +0.119, $p < 0.001$), consistent with the well-characterised metabolic effects of glucocorticoids on hepatic glucose output and peripheral insulin resistance.  The furosemide versus spironolactone comparison---testing opposing potassium effects of two diuretics (loop diuretic causing potassium wasting versus potassium-sparing mineralocorticoid receptor antagonist)---yielded a difference-in-differences of $-$0.061 ($p < 0.001$), a within-class pharmacological distinction that was completely null under all standard pooled methods ($p > 0.3$).  This result suggests the incident-user design has greater sensitivity for detecting directionally consistent pharmacological associations.  Vasopressor use was associated with increased mortality ($\Delta$ +0.069, $p < 0.001$), correctly reflecting confounding by indication in critical illness.

\subsubsection{The four wrong directions}

The 4 incorrect directions fall into interpretable categories that are consistent with the model reproducing observational associations, including the biases inherent in observational data.  The insulin $\to$ glucose result (DiD +0.104, wrong direction) represents irreducible confounding by indication: insulin is prescribed \emph{for} hyperglycaemia, and the model reproduces this observational association.  The anticoagulant class-level INR result (DiD $-$0.011, $p$ = 0.20) is diluted by heterogeneous mechanisms across the anticoagulant class.  The steroid--sepsis mortality association ($\Delta$ +0.014, wrong direction) is consistent with treatment intensity confounding---steroids are given to sicker sepsis patients.  The statin--glucose expected-null violation (DiD +0.012, $p$ = 0.002) suggests residual metabolic confounding, though the effect size is small.

The overall pattern---partial directional recovery under substantial residual confounding---is consistent with a model that has learned the observational associations in its training data without distinguishing correlation from causation.  The four incorrect directions include confounding by indication (insulin), treatment intensity confounding (steroids in sepsis), class-level heterogeneity (anticoagulants), and residual metabolic confounding (statin--glucose).  The pharmacological association testing framework is useful because it reveals both signal and failure, not because the failures validate the method.

\subsubsection{Cross-dataset pooled summary}

Across 26 drug--outcome comparisons on open-access datasets (MIMIC-IV and INSPECT; excluding null controls), FlatASCEND recovered 20 correct pharmacological association directions (\Cref{tab:tte-unique}).  A permutation test (100,000 permutations) yielded $p$ = 0.003, with the null expectation at 13 of 26 correct by chance.  The incident-user analysis reveals, however, that many individually ``correct'' pooled directions have near-null effect sizes---the pooled permutation test provides evidence for collective pharmacological learning, while the incident-user framework provides per-comparison assessment.

\begin{table}[h]
\centering
\caption{Unique pharmacological comparisons by dataset.  Each comparison counted once.  Correct = direction matches published trial evidence.  The incident-user design on MIMIC-IV is the primary analysis.}
\label{tab:tte-unique}
\small
\begin{tabular}{lccccl}
\toprule
\textbf{Dataset} & \textbf{Correct (non-null)} & \textbf{Significant} & \textbf{Expected-null} & \textbf{Total} & \textbf{Method} \\
\midrule
MIMIC-IV (primary) & 6/10 & 9/10 & --- & 10 & Incident-user v2 \\
INSPECT & 5/7 & 0/7 & 1/1 & 8 & Standard \\
\midrule
\textbf{Open-data total} & \textbf{11/17} & \textbf{9/17} & \textbf{1/1} & \textbf{18} & \\
\bottomrule
\end{tabular}
\begin{flushleft}
\footnotesize
Correct and Significant denominators exclude expected-null comparisons.  eICU-CRD was used for mortality and representation evaluation, not pharmacological testing.  A permutation test across 26 comparisons (including additional overlapping MIMIC-IV analysis methods from Table~\ref{tab:tte-methods}) yields $p$ = 0.003.
\end{flushleft}
\end{table}

\begin{table}[h]
\centering
\caption{Method-specific MIMIC-IV analyses (partial overlap with Table~\ref{tab:tte-unique}).  Some drug--outcome pairs are evaluated by multiple methods.  Development-set results shown for architectural context only.}
\label{tab:tte-methods}
\small
\begin{tabular}{llcccc}
\toprule
\textbf{Dataset} & \textbf{Method} & \textbf{Correct} & \textbf{Significant} & \textbf{Expected-null} & \textbf{Total} \\
\midrule
\textit{Dev (Level 1)} & Standard & 2/3 & 2/3 & 2/2 & 5 \\
\textit{Dev (Level 2)} & Standard & 3/3 & 0/3 & 2/2 & 5 \\
\midrule
MIMIC-IV class & Standard & 6/7 & 2/7 & --- & 7 \\
MIMIC-IV drug & Head-to-head & 5/8 & 0/8 & --- & 8 \\
\bottomrule
\end{tabular}
\end{table}

On INSPECT (outpatient, Stanford), 5 of 7 evaluable comparisons recovered the correct direction, including anticoagulant $\to$ INR (OR = 1.080), ACE inhibitor $\to$ potassium elevation (OR = 1.020), and a correctly null metformin versus sulphonylurea comparison.

Head-to-head within-class comparisons on MIMIC-IV recovered the null result of the VASST trial\citep{russell2008vasst} (vasopressin versus norepinephrine, OR = 0.986, $p$ = 0.70), the direction of the PROVE-IT trial\citep{cannon2004proveit} (atorvastatin versus pravastatin, OR = 0.965), and the direction of the COMET trial\citep{poole2003comet} (carvedilol versus metoprolol, OR = 0.916, $p$ = 0.094).

\subsubsection{DPO negative result: reward exploitation on clinical surrogates}
\label{sec:dpo-negative}

Direct preference optimisation\citep{rafailov2024dpo} with a pharmacological reward signal was applied to investigate whether pharmacological association recovery could be improved through training.  This is a documented negative result.  Four training runs with learning rates from $1 \times 10^{-6}$ to $2 \times 10^{-5}$ were conducted.  The best checkpoint destroyed all pre-existing correct associations (3/3 to 0/3).  The strongest comparison---GLP-1 receptor agonist versus sulphonylurea---reversed from OR = 0.98 (correct) to 2.03 (wrong direction, $p < 0.001$).  The model learned to insert low-quintile biomarker tokens directly into treatment-arm trajectories rather than producing pharmacologically plausible trajectories.

The structural cause was circularity: the reward function and evaluation metric both measured surrogate biomarker token distributions.  Preference optimisation found the shortest path---directly manipulating token frequencies---rather than learning the causal mechanisms by which drugs affect biomarkers through physiological pathways.  This finding is directly relevant to any application of reinforcement learning from human feedback to clinical models where the reward and evaluation share an outcome space.

\section{Discussion}
\label{sec:discussion}

\subsection{Generation: addressing multi-step trajectory generation}

FlatASCEND demonstrates multi-step clinical trajectory generation with continuous inter-event time prediction from an autoregressive model trained on electronic health records.  The key innovations are the zero-inflated log-normal time head, which captures the bimodal structure of clinical event timing without discretisation, and free-running scheduled sampling, which exposes the model to its own generation errors during training and raised Jaccard from 0.536 to 0.843 in a single change.  The resulting model achieves Jaccard 0.889--0.954 on the primary trained open-access datasets (MIMIC-IV and INSPECT) with 0\% mode collapse (0.857 on the proprietary development dataset with 2\% collapse).  Zero-shot transfer to eICU-CRD degrades sharply (Jaccard 0.682) due to distributional shift, but brief site adaptation (5,000 steps) recovers Jaccard to 0.820, suggesting that performance is substantially recoverable with brief site adaptation.

The efficiency of this approach is noteworthy.  At 14.5 million parameters, FlatASCEND is two orders of magnitude smaller than NEP (1--8 billion parameters) and one order of magnitude smaller than RAVEN (144 million parameters).  The larger models focus on next-visit prediction or embedding extraction rather than multi-step generation, so direct performance comparison is not possible, but the parameter efficiency is notable.  The flat composite token design---one token per clinical event, no grammar to learn---eliminates the structural validity failures that plagued earlier structured approaches.\citep{sainsbury2025ascendgpt}  The progressive curriculum on the development dataset suggests that architectural refinement alone can improve pharmacological accuracy (2/3 to 3/3 correct directions) without explicit causal training.

The generative evidence is strongest for short-horizon ICU data (MIMIC-IV), where timing calibration, category transitions, and per-patient fidelity are all close to held-out data.  Outpatient generation (INSPECT) is materially weaker: the model under-generates long-gap events, producing trajectories covering days rather than months.  This limitation should be considered when interpreting the generative claim: FlatASCEND demonstrates reliable short-horizon conditional generation, with weaker long-range outpatient temporal fidelity.  The baseline comparison shows that unconditional distributional metrics (Jaccard, timing KS) do not discriminate between learned conditional generation and trivial sampling from the marginal distribution.  The prompt-shuffle ablation provides supporting evidence: destroying clinical context by randomising prompt token order reduced effect sizes by 2.0--2.2$\times$ for both mechanistic controls (steroid $\to$ glucose and diuretic $\to$ potassium), while the confounding-driven insulin $\to$ glucose association was unaffected (0.9$\times$).  This dissociation suggests that the model uses patient-specific clinical context for mechanistic pharmacological associations but not for confounding-driven ones.

\subsection{Pharmacological testing: patient-level paired differences in simulated outcomes}

The transition from pooled trajectory-level analysis to the incident-user design with patient-level Wilcoxon signed-rank tests addresses pseudo-replication, which is the primary source of inflated statistical power in generative pharmacological testing.  The incident-user framework eliminates pseudo-replication (each patient contributes one observation, not 200 synthetic trajectories), controls for pre-treatment outcome levels through within-patient differencing, and produces results that are directly interpretable as patient-level paired differences in simulated outcomes under forced treatment tokens.

Nine of 10 comparisons reach patient-level significance at $N$ = 500, indicating that the model generates systematically different trajectories in response to different drug interventions---a property that requires drug-specific distributional shifts consistent across hundreds of patients.

However, statistical significance is not the same as pharmacological correctness.  Of the 6 correct directions, 4 are mechanistic pharmacological recoveries---warfarin's anticoagulant effect (DiD +0.052), corticosteroid-induced hyperglycaemia (DiD +0.119), and two diuretic $\to$ potassium comparisons---where the model recovers a direct drug--biomarker relationship.  The remaining 2 correct directions are treatment-context associations (vasopressor $\to$ mortality from confounding by indication, statin in sepsis with small-magnitude effect compatible with the SAILS trial).  The 4 incorrect directions are consistent with the model also capturing confounding by indication (insulin prescribed for hyperglycaemia), treatment intensity bias (steroids given to sicker sepsis patients), and residual metabolic confounding (statin--glucose association).

The DPO failure provides a concrete warning for clinical reinforcement learning.  When the reward signal (surrogate biomarker distributions) and the evaluation metric share the same outcome domain, preference optimisation learns the shortest path: directly inserting favourable biomarker tokens into treatment-arm trajectories.  The GLP-1 receptor agonist OR flipped from 0.98 (correct, consistent with the LEADER trial) to 2.03 (wrong direction, $p < 0.001$), a complete reversal of a well-validated pharmacological association.  This is directly relevant to any application of reinforcement learning from human feedback to clinical models where the reward and evaluation share an outcome space.  Breaking this circularity requires that the reward and evaluation operate in different outcome domains.  Three non-circular alternatives merit investigation: (1) training on surrogate biomarker shifts while evaluating against hard diagnostic endpoints (myocardial infarction, stroke, death tokens)---the MIMIC-IV and INSPECT vocabularies now support this; (2) using a discriminative clinical plausibility verifier as reward rather than outcome-based reward; (3) applying classifier-free guidance at inference time to amplify the drug-conditional signal without modifying model weights.

\subsection{Limitations}

\paragraph{ZILN time head and long-range outpatient timing.}  The zero-inflated log-normal time head was designed for the bimodal structure of ICU event timing (many same-visit events at $\dt = 0$, right-skewed inter-visit gaps).  For outpatient data with gaps of weeks to months, the log-normal component may be insufficiently heavy-tailed.  Alternative distributions (e.g., a mixture of log-normals, or a discretised time-bucket approach for outpatient settings) were not evaluated and may improve long-range temporal fidelity.

\paragraph{Trajectory validation has known gaps.}  The timing calibration, conditional dependency, and longitudinal fidelity analyses provide stronger evidence than Jaccard alone, but important gaps remain.  On INSPECT, generated trajectories cover substantially less elapsed time than held-out continuations (median 10 versus 154 days), indicating that the model under-generates long-gap outpatient events.  Per-patient quintile values differ by approximately 1 quintile from the held-out data on average, meaning individual-level biomarker trajectories are approximate rather than precise.  No external clinician adjudication of generated trajectories was performed, and conditional dependency preservation is weaker on outpatient data (Frobenius 0.438) than ICU data (0.083).

\paragraph{Associational, not causal.}  The pharmacological association testing framework recovers observational associations validated against prior pharmacological knowledge.  These are not causal estimates.  Confounding by indication, healthy-user bias, and setting-specific prescribing patterns all influence the results.  The 4 of 10 incorrect directions in the incident-user analysis directly illustrate these limitations.  The mapping between our comparisons and published randomised trials is approximate: GLP-1 receptor agonist versus sulphonylurea is not equivalent to LEADER (liraglutide versus placebo)---the direction is comparable, but the magnitude is not.

\paragraph{Baselines are unconditional.}  The unigram and bigram baselines match or exceed FlatASCEND on aggregate distributional metrics because they sample directly from empirical distributions.  A comparison against a conditional baseline (e.g., a simpler autoregressive model without the ZILN time head or scheduled sampling) would better isolate the contribution of each architectural component.  An ablation study on the open-access datasets was not performed because all architecture decisions were made on the proprietary development dataset.

\paragraph{Pseudo-replication in pooled analyses.}  The 200 trajectories generated per patient per arm are Monte Carlo samples from the model's learned distribution, not biological replicates.  The incident-user design addresses this by aggregating to patient-level statistics, but the pooled cross-dataset permutation test ($p$ = 0.003) should be interpreted as a collective statistic, not as evidence for any individual comparison.

\paragraph{Zero-shot cross-site transfer requires adaptation.}  Under zero-shot transfer to eICU-CRD (208 hospitals), generative quality degraded sharply (Jaccard 0.682, perplexity 776) due to distributional shift in token category proportions (eICU has 12$\times$ more diagnosis tokens than MIMIC-IV).  Brief site adaptation (5,000 steps, $\sim$30 minutes) recovered Jaccard to 0.820 and perplexity to 27.4, suggesting the failure is a correctable distribution-shift problem.  Nevertheless, site-specific adaptation would be required for deployment beyond the training institution.

\paragraph{ICU-dominated open data.}  MIMIC-IV and eICU-CRD are both ICU cohorts.  Outpatient validation relies on a single dataset (INSPECT, 19,000 patients), which limits statistical power for some comparisons and restricts the generalisability of outpatient findings.

\paragraph{Proprietary development dataset.}  The development dataset used for architecture design and the progressive training curriculum cannot be independently reproduced.  The primary results (generation quality, pharmacological testing, trajectory validation) are reported on open-access PhysioNet data, but the architectural choices and the 2/3$\to$3/3 pharmacological improvement were informed by development-set experiments that cannot be independently verified.

\paragraph{Dataset sizes.}  INSPECT is small (19,000 patients) relative to the ICU datasets.  This affects statistical power for pharmacological comparisons at windowed intervals.

\subsection{The pharmacological testing framework}

The two contributions of this paper---generative modelling and pharmacological testing---are not independent findings but components of an analysis approach that could be applied to other generative EHR models.  Generation quality must be established before pharmacological testing is meaningful (unrealistic trajectories would produce meaningless associations).  Pharmacological testing then assesses whether the model's generated trajectories are consistent with clinically relevant structure.

This framework is applicable to any generative EHR model.  Other groups can apply the incident-user framework to their own models to assess pharmacological learning.  The incident-user design, which eliminates pseudo-replication and produces patient-level statistics, is a methodological contribution independent of the specific model.

\section{Methods}
\label{sec:methods}

\subsection{FlatASCEND architecture}
\label{sec:architecture}

FlatASCEND represents each clinical event as a single flat composite token combining clinical category, measure, and value (\eg, \tok{LAB:HBA1C:Q4}, \tok{MED:METFORMIN}, \tok{DX:SEPSIS}).  The vocabulary supports five categories: laboratory results (measure + ordinal quintile), vital signs (measure + ordinal quintile), medications (drug class or individual drug), diagnoses (coded clinical events), and demographics/special tokens (\tok{SEX:F}, \tok{AGE:D5}, \tok{[BOS]}, \tok{[EOS]}, \tok{[DEATH]}, \tok{[PAD]}).  Visit boundaries are encoded implicitly through the time head ($\dt = 0$ for within-visit, $\dt > 0$ for between-visit events), eliminating the need for explicit visit separator tokens.

The model is a GPT-2-style autoregressive transformer with ALiBi\citep{press2022alibi} for position encoding and dual-head output:

\begin{itemize}[leftmargin=2em]
    \item \textbf{Content head} (weight-tied): Linear projection from hidden states to vocabulary logits, with weights tied to the input embedding matrix.
    \item \textbf{Time head} (zero-inflated log-normal): A two-layer MLP producing three parameters:
    \begin{equation}
        (\mathrm{logit}_z, \mu, \log\sigma) = \mathrm{TimeHead}(\mathbf{h}_L^{(t)})
    \end{equation}
    Time deltas are sampled as $\dt = 0$ with probability $\sigma(\mathrm{logit}_z)$, or $\dt \sim \mathrm{LogNormal}(\mu, \sigma)$ otherwise.
\end{itemize}

Input representations combine content and timing:
\begin{equation}
    \mathbf{x}_t = \mathbf{e}_{\text{content}}(\text{token}_t) + \text{TimeMLP}(\log(1 + \dt_t))
\end{equation}

Laboratory and vital sign embeddings are factored into measure identity and ordinal level:
\begin{equation}
    \mathbf{e}_{\text{token}} = \mathbf{e}_{\text{base}}[\text{measure}] + \mathbf{e}_{\text{ordinal}}[\text{bin}]
\end{equation}
where $\mathbf{e}_{\text{base}}$ captures measure identity and $\mathbf{e}_{\text{ordinal}}$ captures ordinal level, creating an inductive bias toward consistent ordinal relationships across measures.

The training loss combines content cross-entropy, time prediction, and ordinal earth mover's distance:
\begin{equation}
    \mathcal{L} = \mathcal{L}_{\text{CE}} + 0.5 \cdot \mathcal{L}_{\text{ZILN}} + 0.25 \cdot \mathcal{L}_{\text{EMD}}
\end{equation}

Category-weighted cross-entropy addresses class imbalance: LAB 2.0$\times$, VITAL 2.0$\times$, MED 1.0$\times$, DX 3.0$\times$, SMOKE 2.0$\times$, AGE 0.25$\times$.  The scaled model configuration (used for all reported results) uses approximately 14.5 million parameters: 384-dimensional hidden states, 8 layers, 12 attention heads, 1536-dimensional feed-forward layers.

\subsection{Training procedure}

All models were trained with AdamW ($\beta_1 = 0.9$, $\beta_2 = 0.999$, weight decay = 0.01), peak learning rate $3 \times 10^{-4}$, linear warmup over 2,000 steps, and cosine decay.  Gradient norms were clipped to 1.0.  Training used mixed-precision (bfloat16 autocast) on a single NVIDIA A100 (80 GB).  Effective batch size was 32 patients per step.  Maximum sequence length was 512 tokens for MIMIC-IV and 1,024 for INSPECT and the development dataset.

\subsubsection{Free-running scheduled sampling}

A single contiguous autoregressive segment replaced approximately 50\% of each training sequence.  The segment start position was randomised uniformly between tokens 2 and 20.  From that position, the model generated autoregressively with temperature 0.8 and top-$k$ = 20, with generated tokens replacing ground truth for the remainder of the sequence.  The loss was computed on all positions including the autoregressive segment.

End-of-sequence and death tokens were suppressed during autoregressive segments by setting their logits to $-\infty$.  Without this suppression, 40--60\% of training segments collapsed to immediate termination within 5,000 steps.  This training strategy was the single largest improvement in generation quality, raising Jaccard from 0.536 to 0.843 in a single version change.

\subsection{Datasets}
\label{sec:dataset-methods}

The development dataset is a proprietary longitudinal diabetes registry (approximately 61,000 patients, 76-token vocabulary after diversity filtering) used exclusively for architecture development and the progressive training curriculum.  Diversity filtering selected patients with at least 100 tokens and 15 unique token types.

INSPECT is a Stanford outpatient dataset in MEDS/OMOP format (approximately 19,000 patients, 220 tokens, median sequence length 615 tokens), including 22 diagnosis codes, 30 LOINC-mapped laboratory measures, and 32 medication classes.

MIMIC-IV\citep{johnson2023mimiciv} version 3.0 comprises 424,974 admissions (303,920 after $\geq$20-token filtering), with 2.6\% in-hospital mortality.  The primary class-level vocabulary (220 tokens) includes 150 laboratory tokens (30 measures $\times$ 5 quintiles), 32 medication classes, 22 diagnosis codes, and 16 demographic/special tokens.  A separate drug-level expansion (6 medication classes to 33 individual drugs, yielding 239 tokens) was used for within-class head-to-head comparisons only; all other analyses use the 220-token class-level model.

eICU-CRD\citep{pollard2018eicu} covers 208 US hospitals (200,859 stays, 5.3\% mortality).  The identical 220-token MIMIC class-level vocabulary was applied with MIMIC-derived quintile boundaries---a deliberate zero-shot protocol with no adaptation to eICU statistics.  Tokens present in eICU but absent from the MIMIC vocabulary (0.7\% of all tokens) were mapped to the padding token.  For the site-adaptation experiment, the MIMIC-IV-trained model was fine-tuned on eICU-CRD training data for 5,000 additional steps with learning rate $1 \times 10^{-4}$ (reduced from the pretraining rate of $3 \times 10^{-4}$), 200-step warmup, and cosine decay.

Patient-level splits (80/10/10 train/validation/test) were used for all datasets.  For MIMIC-IV, splits were at the patient level to prevent information leakage across admissions from the same patient.

\subsection{Generative pharmacological association testing}
\label{sec:tte-methods}

\subsubsection{Standard protocol}

For each comparison: (1) select eligible patients from the validation set (prompt must contain the outcome biomarker type, must not contain the forced drug); (2) append the intervention token with $\dt = 0$ (treatment arm) and the comparator token (control arm); (3) generate $N$ = 200--500 independent trajectories per patient per arm (temperature 1.0, top-$k$ = 20); (4) extract outcomes within the follow-up window (laboratory quintiles for surrogates, \tok{[DEATH]} for mortality); (5) fit a generalised linear model with logit link, treatment indicator, period fixed effects, and cluster-robust standard errors (clustered by patient).  During generation, all same-class medications are suppressed.

\subsubsection{Person-period conversion}

Generated trajectories were divided into fixed-length periods (7 days for MIMIC-IV, 30 days for INSPECT and the development dataset).  Within each period, the outcome was measured as the mean laboratory quintile (surrogate outcomes) or presence of \tok{[DEATH]} (mortality).  This person-period structure enabled the use of period fixed effects that account for time-varying outcome patterns.

\subsubsection{Incident-user design}
\label{sec:incident-user-methods}

The incident-user design is the primary analysis framework, designed to address pseudo-replication and pre-treatment confounding.  For each patient: (1) identify the position of the first occurrence of the intervention drug in the clinical sequence; (2) set the prompt boundary to immediately before this position---the ``must not contain the forced drug'' criterion in the standard protocol applies to the prompt portion, not the full patient history; patients may have received same-class agents earlier in their history, but the prompt boundary is set before the first occurrence of the specific intervention token; (3) record the mean outcome quintile in the last 5 prompt tokens containing the relevant biomarker as the pre-treatment baseline; (4) generate $N$ trajectories in both treatment and control arms, with all same-class medications suppressed during generation to prevent cross-arm contamination---this suppression may create asymmetric clinical plausibility between arms if the comparator drug class is also common in the patient population, which is an inherent limitation of the forced-intervention design; (5) compute within-patient change from baseline in each arm, yielding a patient-level risk difference; (6) test the distribution of patient-level risk differences using the Wilcoxon signed-rank test.  Both arms are model-generated continuations from the same patient prefix, so the design uses within-patient differencing of simulated outcomes to produce one observation per patient.

The incident-user analysis was conducted with $N$ = 500 patients per comparison on MIMIC-IV, providing adequate statistical power for patient-level inference.  A separate power analysis confirmed that $N$ = 200 was underpowered for most comparisons (warfarin $\to$ INR: $p$ = 0.717 at $N$ = 200 versus $p$ = 0.0002 at $N$ = 1,500).

\subsubsection{Permutation test}

The permutation test assessed whether the number of correct association directions significantly exceeded chance.  For each of 100,000 permutations, the binary correct/incorrect labels for the 26 open-data comparisons were randomly shuffled, and the count of ``correct'' labels was recorded.  The null expectation was 13 of 26 (50\%).  The observed count of 20 yielded $p$ = 0.003 (proportion of permutations with $\geq$20 correct).

\subsubsection{Prompt-shuffle ablation}
\label{sec:shuffle-methods}

The prompt-shuffle ablation tests whether patient-specific clinical context affects pharmacological association recovery.  For each patient, the prompt tokens were randomly permuted (a single random permutation per patient), with the first two positions (typically \tok{[BOS]} and a demographic token) preserved to maintain sequence structure.  Token times were permuted jointly with their tokens (each token retained its original time delta).  Prompt lengths were unchanged.  The forced intervention and comparator tokens were appended identically to both ordered and shuffled conditions.

The ablation was conducted on three comparisons (steroid $\to$ glucose and diuretic $\to$ potassium as mechanistic pharmacological controls, insulin $\to$ glucose as a confounding-driven association), with $N$ = 200 patients and 100 samples per arm (versus 500 patients and 200 samples in the main incident-user analysis).  Effect sizes in the ablation are therefore expected to differ from the main analysis due to the smaller sample and different random patient selection.  The analysis was run on the class-level MIMIC-IV model (step 100,000).

\subsection{Statistical analysis}

Pharmacological associations were estimated using two complementary approaches.  The standard protocol used generalised linear models with logit link, treatment indicator, period fixed effects, and cluster-robust standard errors (clustered by patient).  The incident-user design used the Wilcoxon signed-rank test for paired within-patient comparisons at the patient level, with each patient contributing a single difference-in-differences estimate.  The permutation test used 100,000 random label shuffles.  All $p$-values are two-sided.

\section*{AI Technology Disclosure}

Claude (Anthropic) was used for analysis assistance, code review, and manuscript editing during the preparation of this work.  All scientific decisions, experimental design, data analysis, and interpretation were performed by the authors.

\section*{Ethics and Data Access}

Development dataset access was granted under an institutional data governance framework for research use.  MIMIC-IV data was accessed through PhysioNet under the PhysioNet Credentialed Health Data Use Agreement.  INSPECT is available through PhysioNet under the Stanford University data use agreement.  eICU-CRD data was accessed through PhysioNet.  All analyses were conducted on de-identified data.  No individual patient data is presented; all results are population-level aggregate statistics from generated trajectories.

\section*{Code and Data Availability}

The model architecture, training pipeline, and pharmacological association testing code are hosted at \url{https://github.com/csainsbury/ascend-flat} (currently private; contact the corresponding author for collaborator access).  MIMIC-IV, INSPECT, and eICU-CRD data are available through PhysioNet\citep{goldberger2000physionet} with credentialed access.

\clearpage
\section*{Supplementary Material}

\subsection*{Supplementary Table S1: MIMIC-IV incident-user pharmacological association results (full detail)}

\begin{table}[H]
\centering
\small
\begin{tabular}{llccccl}
\toprule
\textbf{Comparison} & \textbf{Type} & \textbf{DiD} & \textbf{95\% CI} & \textbf{Wilcoxon $p$} & \textbf{Dir.} & \textbf{Evidence} \\
\midrule
Warfarin $\to$ INR & Surr. & +0.052 & [+0.038, +0.066] & $<0.001$ & Correct & Vit.\ K inhibition \\
Steroid $\to$ glucose & Surr. & +0.119 & [+0.104, +0.133] & $<0.001$ & Correct & Glucocorticoid effect \\
Furosemide vs spiro $\to$ K & Surr. & $-$0.061 & [$-$0.071, $-$0.050] & $<0.001$ & Correct & Loop vs K-sparing \\
Diuretic $\to$ K & Surr. & $-$0.040 & [$-$0.050, $-$0.030] & $<0.001$ & Correct & Loop hypokalemia \\
Vasopressor $\to$ mortality & Mort. & +0.069 & [+0.062, +0.076] & $<0.001$ & Correct\textsuperscript{a} & Confounding by indication \\
Statin sepsis $\to$ mortality & Mort. & $-$0.004 & [$-$0.005, $-$0.002] & 0.0001 & Correct\textsuperscript{b} & SAILS near-null \\
Insulin $\to$ glucose & Surr. & +0.104 & [+0.092, +0.117] & $<0.001$ & Incorrect & Confounding by indication \\
Anticoagulant $\to$ INR & Surr. & $-$0.011 & [$-$0.029, +0.007] & 0.20 & Incorrect & Class heterogeneity \\
Steroid sepsis $\to$ mortality & Mort. & +0.014 & [+0.011, +0.018] & $<0.001$ & Incorrect & Treatment intensity \\
Statin $\to$ glucose & Surr. & +0.012 & [+0.005, +0.020] & 0.002 & Incorrect & Expected-null violation \\
\bottomrule
\end{tabular}
\begin{flushleft}
\footnotesize
$N$ = 500 patients per comparison (443 for steroid sepsis mortality).  DiD = difference-in-differences in mean quintile (surrogate) or death rate difference (mortality).  95\% CIs from $t$-based estimation on patient-level differences.\\
\textsuperscript{a}Treatment-context association.  \textsuperscript{b}Near-null effect consistent with SAILS trial.
\end{flushleft}
\end{table}

\subsection*{Supplementary Table S2: MIMIC-IV head-to-head within-class comparisons}

\begin{table}[H]
\centering
\small
\begin{tabular}{llccl}
\toprule
\textbf{Comparison} & \textbf{Trial emulated} & \textbf{OR} & \textbf{$p$} & \textbf{Direction} \\
\midrule
Vasopressin vs norepinephrine & VASST & 0.986 & 0.70 & Correct (null) \\
Atorvastatin vs pravastatin & PROVE-IT & 0.965 & --- & Correct \\
Carvedilol vs metoprolol & COMET & 0.916 & 0.094 & Correct \\
Apixaban vs warfarin & ARISTOTLE & --- & --- & --- \\
\bottomrule
\end{tabular}
\begin{flushleft}
\footnotesize
Drug-level 239-token model.  OR = odds ratio from standard pooled analysis.  ARISTOTLE comparison included for completeness but result not evaluable due to insufficient apixaban exposure.
\end{flushleft}
\end{table}

\subsection*{Supplementary Table S3: INSPECT pharmacological association testing results}

\begin{table}[H]
\centering
\small
\begin{tabular}{llccl}
\toprule
\textbf{Comparison} & \textbf{Type} & \textbf{OR} & \textbf{$p$} & \textbf{Direction} \\
\midrule
Anticoagulant $\to$ INR & Surrogate & 1.080 & --- & Correct \\
ACE inhibitor $\to$ potassium & Surrogate & 1.020 & --- & Correct \\
Diuretic $\to$ potassium & Surrogate & --- & --- & Correct \\
Insulin $\to$ glucose & Surrogate & --- & --- & Incorrect \\
Statin $\to$ mortality & Mortality & --- & --- & Correct \\
Metformin vs SU $\to$ glucose & Null & --- & --- & Correct (null) \\
PPI vs PPI & Null & --- & --- & Correct (null) \\
\bottomrule
\end{tabular}
\begin{flushleft}
\footnotesize
Standard pooled analysis on INSPECT (outpatient, 220-token vocabulary).  5 of 7 evaluable non-null comparisons recovered the correct direction; 0 of 7 reached statistical significance at $p < 0.05$.  OR values shown where available.
\end{flushleft}
\end{table}

\subsection*{Supplementary Table S4: DPO negative result (development dataset)}

\begin{table}[H]
\centering
\small
\begin{tabular}{llccc}
\toprule
\textbf{Comparison} & \textbf{Stage} & \textbf{OR} & \textbf{$p$} & \textbf{Direction} \\
\midrule
GLP-1RA vs SU & Level 2 (pre-DPO) & 0.98 & --- & Correct \\
GLP-1RA vs SU & Level 3 (post-DPO) & 2.03 & $<0.001$ & Incorrect \\
Insulin vs SU & Level 2 (pre-DPO) & 0.92 & 0.012 & Correct \\
Insulin vs SU & Level 3 (post-DPO) & --- & --- & Incorrect \\
SGLT2i vs SU & Level 2 (pre-DPO) & 0.96 & --- & Correct \\
SGLT2i vs SU & Level 3 (post-DPO) & --- & --- & Incorrect \\
\bottomrule
\end{tabular}
\begin{flushleft}
\footnotesize
Proprietary development dataset (76-token vocabulary, 61K patients).  DPO with surrogate biomarker reward destroyed all 3 pre-existing correct directions.  Level 2 = factored embeddings + EMD loss; Level 3 = Level 2 + DPO.
\end{flushleft}
\end{table}

\subsection*{Supplementary Figure S1: Timing calibration}

Observed versus generated inter-event time distributions for MIMIC-IV and INSPECT.  Log-space histograms, quantile-quantile plots, and cumulative distribution functions for non-zero inter-event times.  Bhattacharyya coefficient 0.982 (MIMIC-IV) and 0.993 (INSPECT) in log-space.

\begin{figure}[h]
\centering
\includegraphics[width=\textwidth]{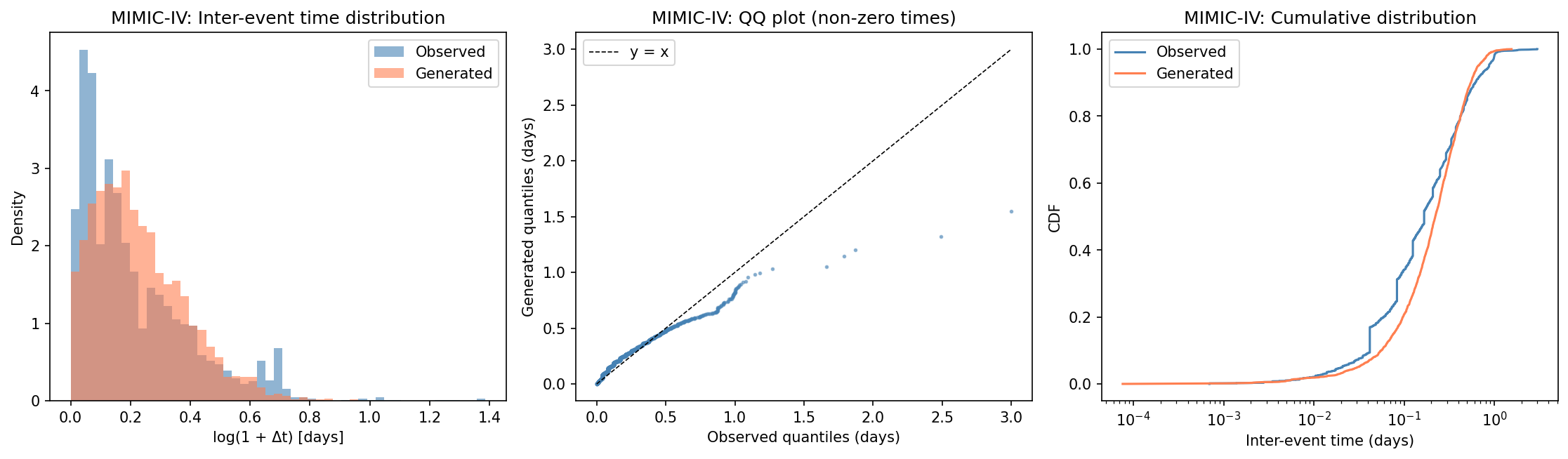}
\caption*{Supplementary Figure S1a: MIMIC-IV timing calibration.}
\end{figure}

\begin{figure}[h]
\centering
\includegraphics[width=\textwidth]{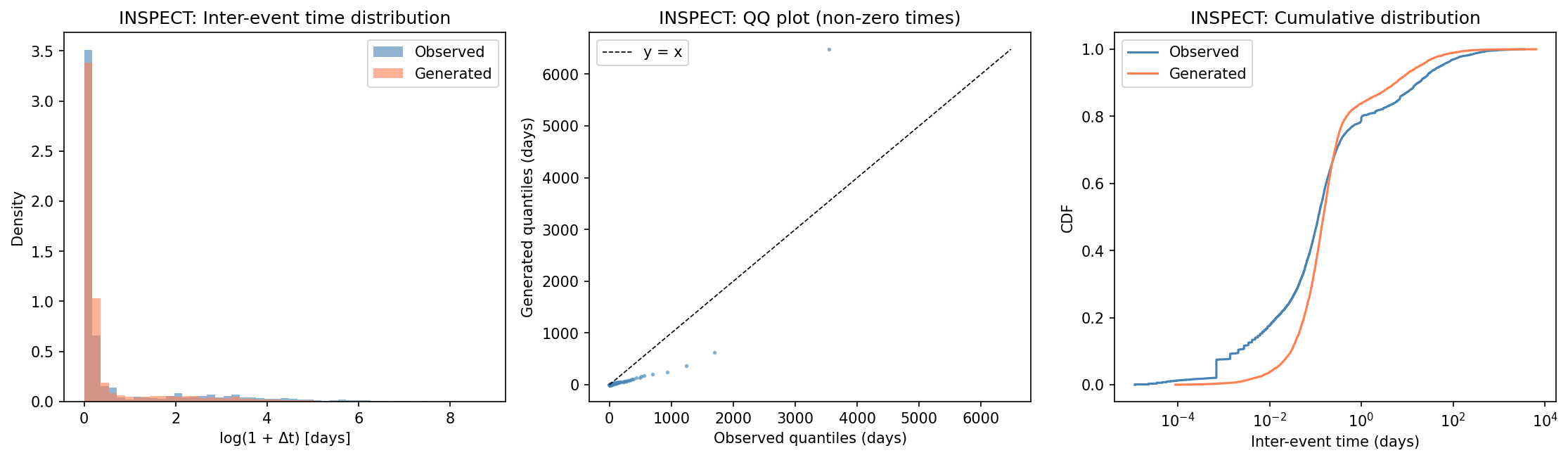}
\caption*{Supplementary Figure S1b: INSPECT timing calibration.}
\end{figure}


\bibliographystyle{plainnat}

\end{document}